\def\year{2017}\relax
\documentclass[letterpaper]{article}
\usepackage{aaai17}
\usepackage{times}
\usepackage{helvet}
\usepackage{courier}
\usepackage{graphicx}
\usepackage{wrapfig}
\usepackage{url}
\usepackage{amsmath}
\usepackage[font={small}]{caption}
\usepackage{color}
\usepackage{amsmath,amssymb}
\usepackage{enumitem}
\usepackage{bm}

\newcommand\numberthis{\addtocounter{equation}{1}\tag{\theequation}}

\begin{document}
%
\title{A Study of Question Effectiveness Using Reddit ``Ask Me Anything'' Threads}
\author{
Kristjan Arumae, Guo-Jun Qi, Fei Liu\\
University of Central Florida, 4000 Central Florida Blvd., Orlando, Florida 32816\\
{\tt k.arumae@knights.ucf.edu, guojun.qi@ucf.edu, feiliu@cs.ucf.edu}
}

\maketitle

\begin{abstract}

Asking effective questions is a powerful social skill.
In this paper we seek to build computational models that learn to discriminate effective questions from ineffective ones.
Armed with such a capability, future advanced systems can evaluate the quality of questions and provide suggestions for effective question wording.
We create a large-scale, real-world dataset that contains over 400,000 questions collected from Reddit ``Ask Me Anything'' threads. 
Each thread resembles an online press conference where questions compete with each other for attention from the host.
This dataset enables the development of a class of computational models for predicting whether a question will be answered.
We develop a new convolutional neural network architecture with variable-length context and demonstrate the efficacy of the model by comparing it with state-of-the-art baselines and human judges.


\end{abstract}

\section{Introduction}

Learning to ask effective questions is important in many scenarios.
For example, doctors are trained to ask effective questions to gather necessary information from patients in a short time~\cite{Molla:2007};
journalists frame their questions carefully to elicit answers~\cite{Vlachos:2014}; 
students post their questions on class discussion forums to seek help on assignments~\cite{Moon:2014}.
Naturally the more effective a question, the better the information-seeking or problem-solving purpose is served.
Despite its importance, the problem of ``what constitutes a good question'' is largely unexplored.
This paper seeks to close the gap by designing and evaluating algorithms that learn to discriminate effective questions---questions that elicit answers---from those that do not.
We conduct a large-scale data-driven study using Reddit ``Ask Me Anything'' (henceforth ``AMA'') discussion threads.

\begin{figure}[t]
\centering
\includegraphics[width=3.2in]{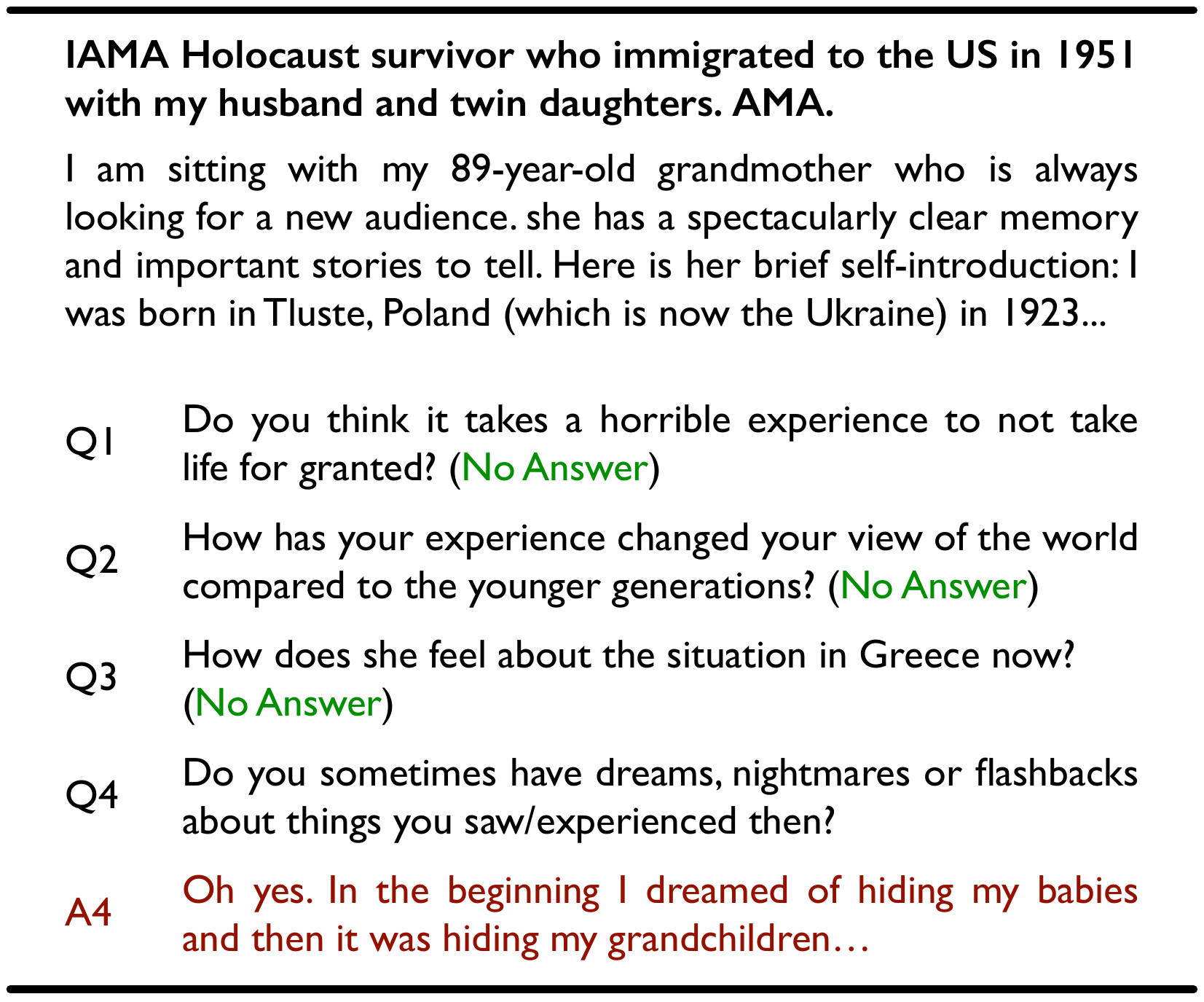}
\caption{A snippet of the AMA thread. It includes the title (shown in bold), a brief intro from the AMA host, four questions (Q1--Q4) posted by Reddit users, among which only the last question was answered by the host (A4).}
\label{fig:example}
\vspace{-0.2in}
\end{figure}

Reddit is a vibrant internet community with more than 36 million registered users~\cite{Reddit:2015}.
It was founded on June 23rd, 2005 and is currently ranked as the 24th most frequently visited website. 
The vast stream of data provides an unprecedented opportunity to study question effectiveness.
Our focus of this work is on the ``Ask Me Anything'' subreddit.
It is a popular subforum with more than 15 million subscribers.
Each AMA thread emulates an online press conference.
An AMA host initiates a discussion thread with ``Ask Me Anything'' or ``Ask Me Almost/Absolutely Anything.''
The host will provide a brief background description and invite others to ask any questions about any topic.
An example is illustrated in Figure~\ref{fig:example}.
The questions in an AMA thread compete with each other for attention from the host. 
The host will chooses to answer some questions while leaving others behind.

In this paper we will investigate whether it is possible to automatically predict which questions will be answered by the host.
We hypothesize that the question wording is important. 
For example, ``\emph{How does she feel about...}'' (Figure~\ref{fig:example}) appears to be a difficult question that requires a delicate answer. 
Further, some topics are considered more favorable/unpleasant than others by AMA hosts. 
We draw on recent development of deep neural networks to predict question answerability.  
The advantage of neural models is the ability to automatically derive question representations that encode both syntactic structure and semantic knowledge.
We develop a novel convolutional neural network (CNN) architecture that considers \emph{variable-length context}.
We draw an analogy between linear classifiers with n-gram features and the new CNN model, and demonstrate improved results.
The contribution of this work includes:
\begin{itemize}
\item A large-scale, real-world question-answering dataset collected from Reddit.\footnote{The Reddit question-answering dataset is available at: \url{http://www.nlp.cs.ucf.edu/downloads/}} 
The dataset contains over 10 million posts and 400,000 questions generated by Reddit users.
It is a very valuable resource for future research on question answering and question suggestion.

\item We introduce a new variable-length context convolutional neural network model and demonstrate its efficacy against state-of-the-art baselines.
We additionally present a human evaluation study to gain further insight into question answerability. 
\end{itemize}

\section{Related Work}


We discuss related work on community question answering and Reddit-inspired natural language processing studies.

A domain of research that is related to this study is community question answering (CQA), where users post questions on discussion forums and solicit answers from the community.
Popular CQA websites include Yahoo! Answers, Ask Ubuntu, Stack Exchange, and Quora.
The question-answering patterns in these websites are different from those of the Reddit AMA.
For example, AMA questions are largely opinion-eliciting and they can be ``any questions about any topic,'' whereas CQA questions are problem-solving oriented and focus on technical topics. 
Multiple factors can contribute to unanswered questions on CQA websites, including the posting time, question quality, user reputation, and reward mechanisms~\cite{Anderson:2012,Li:2012,Liu:2013:EMNLP,Ravi:2014,Nakov:2016}.
CQA corresponds to a single-inquirer multiple-responders setting, whereas AMA corresponds to a multiple-inquirers single-responder setting, because the AMA host single-handedly responds to all questions posted to the discussion thread.
Whether a question will be answered depends on the question content---if it generates sufficient interest that warrants an answer from the host.
Danish, Dahiya, and Talukdar~\shortcite{Danish:2016} analyze a range of factors that influence the AMA question answerability.
In contrast to their work, we focus on developing neural networks that automatically learn question representations that incorporate syntactic and semantic knowledge.

Reddit has been explored for studying various language-related problems.
Bendersky and Smith~\shortcite{Bendersky:2012} investigate characteristics of ``quotable'' phrases.
The machine-detected phrases are quotable if they obtain endorsement from Reddit users.
Wallace et al.~\shortcite{Wallace:2014} leverage contextual information for irony detection on Reddit. 
Schrading et al.~\shortcite{Schrading:2015} develop classifiers to predict texts that are indicative of domestic violence relationships.
Jaech et al.~\shortcite{Jaech:2015} introduce a comment ranking task that predicts the popularity of comments based on language features such as informativeness and relevance.
Ouyang and McKeown~\shortcite{Ouyang:2015} create a corpus of personal narratives using Reddit data. 
Most recently, Tan et al.~\shortcite{Tan:2016} and Wei et al.~\shortcite{Wei:2016} respectively present studies on understanding the mechanisms behind ``persuasion.''
They acquire discussion threads from the ``ChangeMyView'' subreddit. 
Users of this subforum state their views on certain topics, invite others to challenge their views, and finally indicate if the discussion has successfully altered their opinions. 
The studies find that both interaction patterns and language cues are predictive of persuasiveness.
While Reddit has become a rising platform for language-related studies, the Reddit AMA corpus described next is expected to drive forward research in question-answering and question answerability.

\begin{table*}
\setlength{\tabcolsep}{6pt}
\renewcommand{\arraystretch}{1.1}
\centering
\begin{small}
\begin{tabular}{ l | r | r | r | r | r | r | r | r | r }
& \textbf{2009} & \textbf{2010} & \textbf{2011} & \textbf{2012} & \textbf{2013} & \textbf{2014} & \textbf{2015} & \textbf{2016} & \textbf{Total/Avg}\\
\hline
\hline
{\# of threads} & 5,055 & 11,987 & 21,381 & 11,555 & 7,517 & 5,514 & 4,054 & 660 & 67,723\\
{Avg \# of posts per thread} & 87.55 & 97.04 & 94.12 & 214.08 & 308.35 & 395.46 & 338.98 & 447.02 & 247.83\\
{Avg \# of sentences per post} & 3.19 & 3.03 & 2.87 & 2.79 & 2.62 & 2.57 & 2.65 & 2.65 & 2.80\\ 
{Avg \# of words per post} & 49.82 & 46.44 & 43.49 & 41.98 & 38.07 & 37.02 & 39.02 & 40.37 & 42.03\\ 
\hline
\hline
{\# of question posts} & 697 & 2689 & 27,835 & 90,522 & 109,969 & 109,569 & 62,874 & 16,064 & 420,219\\
{\% of question posts w/ answer} & 44.76 & 30.83 & 30.03 & 20.48 & 19.66 & 19.77 & 22.02 & 18.56 & 25.76\\
\end{tabular}
\end{small}
\caption{Statistics of the Reddit AMA corpus. The data ranges from the beginning of Reddit (May 2009) to the end of the data collection period (April 2016). Macro-average scores over all years are reported in the final column.}
\label{tab:dataset}
\vspace{-0.15in}
\end{table*}

\section{The Reddit AMA Corpus}

We describe a methodology for creating a large-scale real-world dataset from Reddit ``Ask Me Anything'' subforum.
Our goal is to collect \emph{all} AMA threads and their associated posts during a span of eight years.
The process is non-trivial, considering the sheer volume of data.
In particular, we are faced with two challenges.
First, the Reddit API returns at most 1000 threads per search query.
To collect all threads, we implement a 30-minute query window to retrieve threads during each time period.
Second, a thread is represented as a JSON object, but certain posts may be missing.
The JSON object is restricted to contain up to 1,500 posts in order to avoid excessive page loading time, whereas the largest AMA threads can reach over 20,000 posts.
To build a complete thread structure, we make additional API calls to retrieve missing posts and insert them back to the thread structure.
An AMA thread contains a collection of posts organized into a tree structure.
Each tree node is a post that includes a variety of useful information, such as author, body text, creation time, and replying posts.

The dataset contains a total of 67,723 discussion threads and over 10 million posts.
Each thread contains 248 posts on average.
The dataset spans eight years, from the beginning of Reddit AMA (May 2009) to end of the data collection period (April 2016).
A number of celebrities have held AMA sessions during this time, including Barack Obama, Arnold Schwarzenegger, Astronaut Chris Hadfield, and Bill Nye.
Other community members with unique experiences are also asked to host AMA sessions.
The dataset is near-complete, missing only 250 threads (0.4\% of all threads)---mostly due to server timeouts.  
More data statistics are presented in Table~\ref{tab:dataset}.
The AMA dataset provides a highly valuable resource for future research on Reddit and question-answering.

\textbf{Preprocessing.}
The goal of preprocessing is to create a collection of question posts and their associated labels. 
If the question post is replied by the AMA host, it is assigned a label of 1, otherwise 0.
We perform a series of thread-level and post-level filtering operations on the dataset to achieve this goal.
First, threads that contain fewer than 100 first-tier posts are removed.
These threads may contain spam or be of low-quality.
Next, the ``AMA Request'' threads are ignored. These threads request certain AMA sessions to be held but are not real AMA discussions.
On the post-level, we restrict the question posts to be of first-tier and within the AMA host's active period.
The active period is defined as the beginning of the thread to the last post of the host.
This setting ensures that the questions are directed towards the host, and the host is aware of the question posts.
We further require question posts to contain a single sentence ending with a question mark.
This allows us to focus on the question content and remove factors such as question length (a few words vs. several paragraphs), multiple questions in a single post, follow-up questions, and non-question content.
The above filtering process generates 420,219 questions. Among them, about 26\% are answered by AMA hosts (see Table~\ref{tab:dataset}).

\section{Our Approach}

So far we have described a large-scale real-world dataset for predicting question effectiveness, we proceed by introducing our proposed solutions. 
We define \textit{effective} questions as those that successfully elicit responses from AMA hosts in a time competitive environment\footnote{Throughout the paper we use question \textit{effectiveness} and \textit{answerability} interchangeably.}.
The task naturally lends itself to a classification formulation where questions that receive answers from AMA hosts are labeled as 1, otherwise 0.
We do not make use of the answer text in this study.
The model learns to discriminate effective questions from ineffective ones based on question body.  
We introduce a new convolutional neural network architecture for this study.
Deep neural models have demonstrated success on a range of natural language processing tasks, most notably machine translation~\cite{Sutskever:2014,Luong:2016}, language generation~\cite{Rush:2015}, and image captioning~\cite{Xu:2015}.
This study presents a novel, context-aware convolutional neural network (CNN) architecture that considers both \textit{context length} and \textit{context variety}.

Concretely, let $\bm{x} = \{x_1, x_2, \cdots, x_\textsf{N}\}$ be a single-sentence question post consisting of $\textsf{N}$ word tokens, where $\textsf{N}$ is the sequence length.
If a question contains more than $\textsf{N}$ words, it is truncated; otherwise, zeros are padded to the end of the word sequence.
Each word is replaced by a word embedding ($\bm{x}_i \in \mathbb{R}^d$) before it is fed to the CNN model.
With a slight abuse of notation, we use $x_i$ to represent the word index in the vocabulary and $\bm{x}_i$ (bold-face) to represent its embedding.
We use the 300-dimension ($d$=300) word2vec embeddings pre-trained on the Google News dataset with about 100-billion words.\footnote{https://code.google.com/archive/p/word2vec/\\
Words that do not have embeddings are set to all zeros. The out-of-vocabulary rate for this dataset is about 2\%.
We additionally trained embeddings using Reddit dataset but they did not yield improved performance over the Google word2vec embeddings.
}

An important building block of the CNN model is the convolutional layer.
It takes a word span as input, concatenates the word embeddings (denoted as $\bm{x}_{j-n+1:j}$ for position $j$ and a span of $n$ words, see Eq.(1)), applies a \textbf{filter} function $h_j=f(\bm{w}^\top \bm{x}_{j-n+1:j} + b)$, and produces a scalar-valued feature representation $h_j$ for the word span (Eq.(2)).
The filter is expected to activate when the word span encodes certain feature; it thus ``memorizes'' the feature using the parameters $\bm{w}$ and $b$.
The convolution process applies each filter to a sliding window of $n$ words that runs over the input sequence.
This process produces a \textbf{feature map} for each filter.
A max-over-time pooling is applied to each feature map to select the most prominent feature $\hat{h}$ in the input sequence (Eq.(3)).
\begin{align*}
& \bm{x}_{j-n+1:j} = [\bm{x}_{j-n+1} ; \bm{x}_{j-n+2} ; \cdots ; \bm{x}_{j}] \numberthis\\
& h_j = f(\bm{w}^\top \bm{x}_{j-n+1:j} + b) \numberthis\\
& \hat{h} = \max_{j=1:\textsf{N}-n+1} h_j \numberthis
\end{align*}

In the aforementioned process, the word span can be of varying sizes ($n$={1,2,3,4,5,$\cdots$}).
Multiple filters can be applied to word span of each size.
The combination of the two factors empowers the model to capture a flexible, enriched representation of the question body.
But, how can we search for the optimal configuration of word span sizes (i.e., \textbf{context length}) and the number of filters applied to word span of each size (i.e., \textbf{context variety}), in order to best depict the prominent features encoded in the word span?

Existing studies use brute force to search for optimal configurations.
Kim~\shortcite{Kim:2014} explores window sizes of \{3,4,5\} words, each with 100 filters.
They are combined in a CNN architecture with one convolution/max-pooling layer, followed by a softmax output layer. 
Albeit simple, the CNN architecture outperforms several state-of-the-art neural models~\cite{Socher:2013,Le:2014}.
Lei, Barzilay, and Jaakkola~\shortcite{Lei:2015} exploit non-consecutive word spans, using tensor decomposition to recognize patterns with intervening words.
They explore spans of \{2,3\} words and \{50,100,200\} filters per word span.
Zhang and Wallace~\shortcite{Zhang:2015} experiment with various combinations of window sizes and numbers of filters per window size.
Their results suggest that searching over the range of 100 to 600 for optimal number of filters is reasonable.

Different from previous studies, we draw an analogy between linear classifiers with n-gram features and convolutional neural networks, and use it to derive the optimal architecture configuration. 
We hypothesize that greater context leads to a higher degree of variety.
Take n-grams for an example, trigrams in theory lead to $O(|\mathcal{V}|^3)$ number of varieties, where $|\mathcal{V}|$ is the vocabulary size.
The intuition behind it is that \emph{the number of filters assigned to a word span correlates with the number of effective features} that can be encoded by a span of $n$ words.

\begin{figure}[t]
\centering
\includegraphics[width=3.3in]{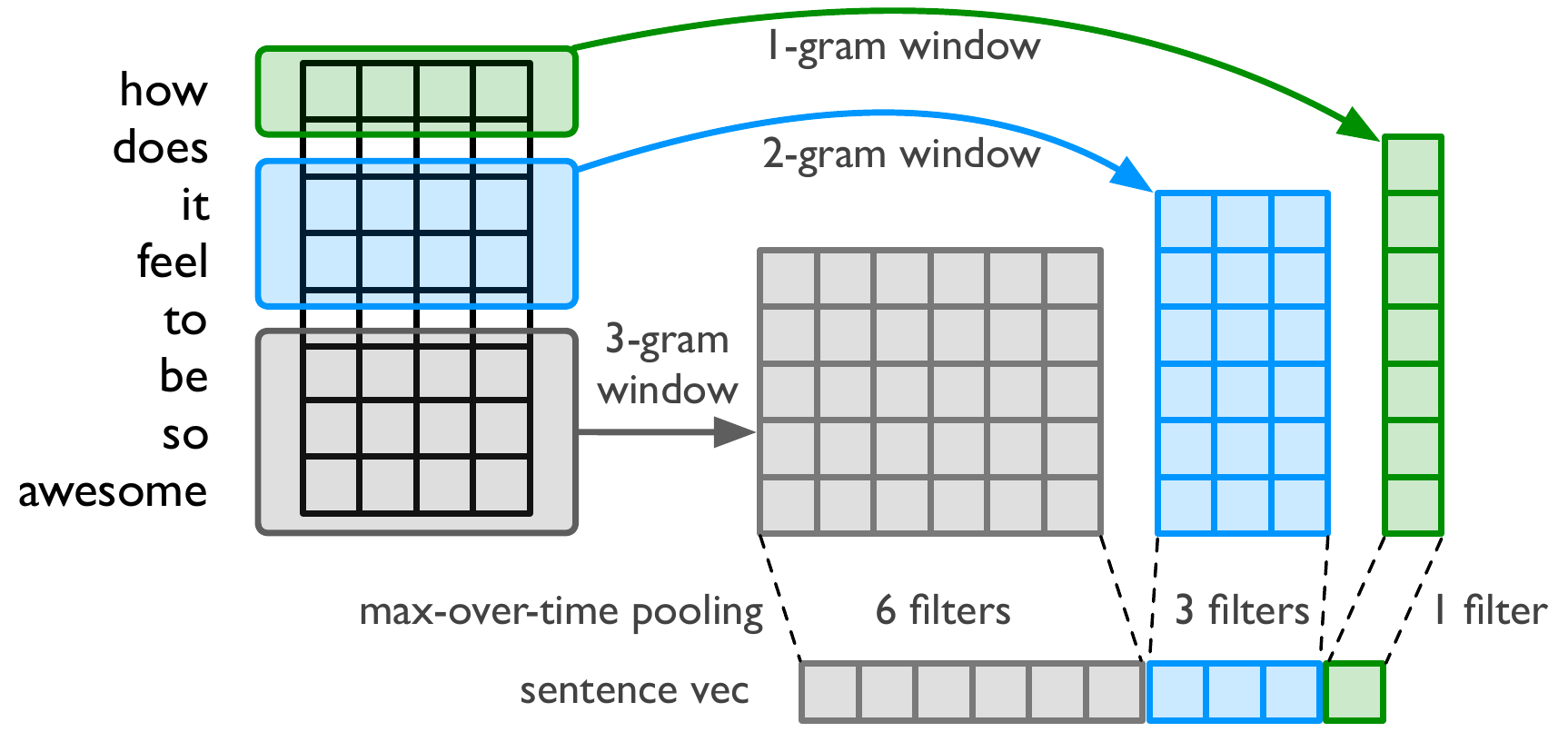}
\caption{A toy example illustrating a convolutional neural network architecture that combines context window of \{1,2,3\}-grams and applies a varying number of filters to each window size.}
\label{fig:toy}
\vspace{-0.15in}
\end{figure}
To this end, we explore n-gram statistics of large-scale, real-world datasets (Table~\ref{tab:ngrams}), including the Europarl corpus~\cite{Koehn:2005} and Google N-grams dataset~\cite{Brants:2006}.
The Europarl corpus contains the European Parliament proceedings collected since 1996.
In this dataset, the number of bigrams is about 15 times the number of unigrams (denoted by $\uparrow_{u}$ 15x),
whereas the number of 3-grams, 4-grams, and 5-grams are of similar sizes, corresponding to averagely 2.4x of the bigram number (denoted by $\uparrow_{b}$ 2.4x).
This is not Zipf's law since we count the number of unique n-grams---regardless of their frequencies---in order to induce a relationship between the window sizes and numbers of effective features.
Note that n-grams that appear only once are filtered out from the Europarl corpus since they are unlikely to be effective features.
In a much larger Google N-grams dataset~\cite{Brants:2006} where n-grams appear less than 40 times are filtered out, similar ratios hold.

To summarize, because a grid search on the optimal combination of window sizes and number of filters per window size is prohibitively expensive, we rely on heuristics derived from n-gram statistics to combine five window sizes of \{1,2,3,4,5\} words and set the numbers of filters to be \{$x$, 20$x$, 60$x$, 60$x$, 60$x$\} respectively, where $x$ is a parameter tunable according to the available computing power.
The ratios are obtained by averaging across the two corpora.
We advocate for a convolutional neural network architecture that combines multiple window sizes and applies a varying numbers of filters per window size.
The approach brings together two considerations on both the context length and context variety, where greater context leads to higher variety. 
Finally, the convolution/max-pooling layer produces a vector representation for each question (illustrated in Figure~\ref{fig:toy}).
We apply a softmax layer on the top to predict whether the question will be answered by the AMA host.



\begin{table}[t]
\setlength{\tabcolsep}{5pt}
\renewcommand{\arraystretch}{1.1}
\centering
\begin{small}
\begin{tabular}{l|r|c|r|c}
& \multicolumn{2}{c|}{\textbf{Europarl Corpus}} & \multicolumn{2}{|c}{\textbf{Google N-grams}}\\
Order & \#Ngrams & Increase & \multicolumn{1}{|c|}{\#Ngrams} & Increase\\
\hline
\hline
1-gram & 53,253 & --- & 13,588,391 & --- \\
2-gram & 816,091 & $\uparrow_{u}$ 15x & 314,843,401 & $\uparrow_{u}$ 23x\\
3-gram & 2,070,512 & $\uparrow_{b}$ 2.5x & 977,069,902 & $\uparrow_{b}$ 3.1x\\
4-gram & 2,222,226 & $\uparrow_{b}$ 2.7x & 1,313,818,354 & $\uparrow_{b}$ 4.2x\\
5-gram & 1,557,598 & $\uparrow_{b}$ 1.9x & 1,176,470,663 & $\uparrow_{b}$ 3.7x\\
\end{tabular}
\end{small}
\caption{N-gram statistics of two large-scale, real-world datasets. Unique n-grams are counted; low-frequency ones are discarded.}
\label{tab:ngrams}
\vspace{-0.1in}
\end{table}

\section{Experiments}

Having described a context-aware CNN architecture in the previous section, we next compare it to a range of baselines and report performance on the Reddit AMA dataset.

\subsection{Baselines and Evaluation Metric}
Our goal is to design and evaluate algorithms that learn to discriminate effective questions from ineffective ones.
The task naturally lends itself to a text classification setting.
We compare the context-aware CNN approach with two baselines.
The first baseline is a logistic regression classifier with bag-of-words features.
We use the LIBLINEAR implementation with $l_2$ regularization~\cite{REF08a}.
The model contains over 100k unigram features with stopwords removed.
A bigram logistic regression model would be computationally prohibitive for this task due to the enormous feature space.
We use the logistic regression baseline because of its proven track record on text classification. 
The same approach has been adopted in a prior work on classifying Reddit posts~\cite{Danish:2016}.

The second baseline is a state-of-the-art convolutional neural network architecture inspired by~\cite{Kim:2014}.
It has a flat structure that contains one convolution layer followed by a max-over-time pooling layer to build the sentence representation.
We use a trigram context window and set the number of filters to be 100. 
This setting has been shown to perform robustly in previous work~\cite{Kim:2014,Lei:2015}.
Because the dataset is imbalanced---only about 25\% of the questions are answered---we create a balanced dataset during training by setting the weights of answered/unanswered questions to be 4/1. 
This applies to all CNN trainings and has shown success on a number of classification tasks~\cite{Huang:2016}.

We create the train/valid/test splits by randomly sampling 100k/10k/100k questions from the 420,219 question collection. 
The train/valid/test data are temporally uniform, meaning there is an even representation of data instances from the entire timeline in each split.
The metric used for model evaluation is the area under the ROC curve (AUC).
Higher AUC scores are better.
When the dataset is highly imbalanced, AUC scores have been shown to be a more appropriate metric than F-scores~\cite{Murphy:2012}.





\subsection{System Evaluation Results}

We report the performance of three systems in Table~\ref{tab:results_all}.
Because the neural network approaches demonstrate performance variation due to the randomness in parameter initiation, 5 independent runs are performed for each test condition. Their mean, min, and max are reported. 

\begin{table}[t]
\setlength{\tabcolsep}{7pt}
\renewcommand{\arraystretch}{1.1}
\centering
\begin{small}
\begin{tabular}{l|c|c|c}
\textbf{System} & \textbf{Mean} & \textbf{Min} & \textbf{Max}\\
\hline
\hline
Logistic Regression & \textbf{0.512} & 0.512 & 0.512\\
Baseline CNN & \textbf{0.516} & 0.509 & 0.525\\
Context-aware CNN ($x$=5) & \textbf{0.523} & 0.503 & 0.537\\
\end{tabular}
\end{small}
\caption{Results on 100k test set as evaluated by AUC scores.}
\label{tab:results_all}
\vspace{-0.15in}
\end{table}

Overall, we found that the context-aware CNN model to produce improved performance than logistic regression and the basic CNN.
Predicting whether or not a question will be answered by the AMA host can be a challenging task.
This is illustrated by the AUC scores of all systems.
There could be multiple reasons. 
First, the three systems employed in this study make extensive use of text features. 
While it allows us to push to the extreme of what can be achieved using text-only features,
other useful non-text information could be incorporated in the future, such as quetion posting time.
Second, each AMA host may have personalize preference on what types of questions they would prefer to answer.
Some AMA hosts stay longer and answer all questions, while others, particularly celebrity hosts, may only answer questions sparsely for a few hours.
Since this is a first and preliminary work on predicting question effectiveness,
we plan to incorporate these factors in future studies.
  
Because the baseline CNN considers an unoptimized combination of context length (window size) and context variety (number of filters), we are curious to know how the two factors individually affect its efficacy in predicting question effectiveness. 
A set of experiments are conducted with varing window sizes of \{1, 2, 3, 4, 5\}-gram and number of filters of \{5, 100, 300\}.
Results are reported on the 10k validation set.
Each condition is again evaluated using 5 runs and the mean scores are reported in Table~\ref{tab:results_context_size}.
When there are only 5 filters (first row), the best performance was achieved using a context window of 4 words.
It appears that the neural model may attempt to memorize a few cue phrases that are particularly good for making predictions.
When there are 300 filters, the model yields the best performance using a window of 1-word.
It seems the network relies on a combination of unigrams to make the prediction.
The context-aware CNN model in this study combines five sub-optimal baseline CNN settings, including (1-gram, 5), (2-gram, 100), (3-gram, 300), (4-gram, 300), and (5-gram, 300) and yields a better performance of 0.537 on the same dataset.

\begin{table}[h]
\setlength{\tabcolsep}{6pt}
\renewcommand{\arraystretch}{1.1}
\centering
\begin{small}
\begin{tabular}{l|c|c|c|c|c}
\textbf{Filters} & \textbf{1-gram} & \textbf{2-gram} & \textbf{3-gram} & \textbf{4-gram} & \textbf{5-gram}\\
\hline
\hline
5 & 0.512 & 0.510 & 0.510 & \textbf{0.515} & 0.511\\
100 & 0.518 & 0.516 & 0.510 & 0.519 & \textbf{0.520}\\
300 & \textbf{0.525} & 0.517 & 0.515 & 0.517 & 0.507\\
\end{tabular}
\end{small}
\caption{Results on 10k valid set. Baseline CNN performance with different window sizes and varying numbers of filters per window.}
\label{tab:results_context_size}
\vspace{-0.1in}
\end{table}

We further explore the effect of different training data sizes. Results are reported in Table~\ref{tab:results_train_size}. 
When the training data is of small size (1$\sim$2k), the context-aware CNN model shows signs of instability. With more data, the performance steadily improves.

\begin{table}[t]
\setlength{\tabcolsep}{7pt}
\renewcommand{\arraystretch}{1.1}
\centering
\begin{small}
\begin{tabular}{l|c|c|c}
\textbf{Train Size} & \textbf{Mean} & \textbf{Min} & \textbf{Max}\\
\hline
\hline
1k & \textbf{0.521} & 0.507 & 0.530 \\
2k & \textbf{0.511} & 0.501 & 0.518\\
5k & \textbf{0.519} & 0.507 & 0.529\\
10k & \textbf{0.523} & 0.508 & 0.542\\
20k & \textbf{0.521} & 0.512 & 0.528\\
50k & \textbf{0.525} & 0.515 & 0.535\\
100k & \textbf{0.537} & 0.531 & 0.544\\
\end{tabular}
\end{small}
\caption{Results on 10k valid set using varied training data size. }
\label{tab:results_train_size}
\vspace{-0.15in}
\end{table}

\subsection{Human Evaluation Results}


Finally, we examine how well humans can distinguish if a question is more effective than another.
The hypotheses we investigate include (1) Do human annotators reach consensus on if certain questions are more likely to be answered than others? (2) Do questions preferred by human annotators agree with those answered by AMA hosts? (3) Will system prediction results be more close to human annotators or AMA hosts?

To answer these questions, we select 300 pairs of questions from our AMA dataset. 
In each pair, the question post A and B come from the same thread.
They are paired with one having received an answer, while the other remains unanswered.
They are temporally adjacent to each other so that the AMA host has a chance to see both questions in a small timeframe---the host receives a notification when new content is posted to the thread---but pick just one to respond to.
We use the Amazon Mechanical Turk (\url{mturk.com}) for this study. 
In each HIT (Human Intelligence Task), a turker is presented with 3 pairs of questions collected from an AMA thread. One question of each pair is answered by the host.
The question order is randomized so that the turkers will not find any specific pattern.
The turkers are also provided with the thread title and background information from the AMA host.
5 turkers are recruited and they are rewarded \$0.05 for completing each HIT.
We request the turkers to be in the U.S. and have a HIT approval rate of 90\% or more.

We observe that the turkers demonstrate a reasonable degree of consensus among themselves: 63\% of the selections are agreed by 4 turkers or more; 30\% of the selections are agreed by all 5 turkers.
On the other hand, there appears to be some discrepancy between the questions chosen to answer by turkers vs. AMA hosts.
We adopt the majority-vote selection from the turkers.
Overall, turkers have agreed with AMA hosts on 55\% of the pairs regarding whether question A or B will be answered.
The context-aware CNN has slightly higher agreement with turkers (54.33\%) than AMA hosts (49.33\%).
Additionally, we hired an undergraduate student to perform the task. The student yields a higher level of agreement with turkers (68.6\%) than AMA hosts (55\%).
These findings suggest that the answered questions are the preference of AMA hosts.
Considering the subreddit has served as an informational as well as promotional platform for the hosts, the AMA hosts may have chosen questions in order to promote personal opinions and these questions may not be the ones that audience would like to have answered.
The human evaluation allows us to derive insights on why some questions are preferred over others, these findings will be incorporated in future work.



\section{Conclusion}
This paper seeks to build computational models that learn to discriminate ``answerable'' questions from those that are not. 
We construct a large-scale question answering dataset that contains over 10 million posts and 400,000 single-sentence questions collected from the Reddit AMA subforum. 
As a preliminary study we present a novel context-aware convolutional neural network architecture that considers both context length and context variety for predicting question answerability. 
While other model architectures such as recurrent neural networks can be explored in the future,
this study presents a first step toward using data-driven approaches for studying question effectiveness.

\bibliography{ref,reddit}
\bibliographystyle{aaai}

\end{document}